\documentclass[10pt, a4paper]{article}

\usepackage[final]{lrec2026} 
\usepackage{times}
\usepackage{latexsym}
\usepackage{tabularx}
\usepackage{graphicx}
\usepackage{booktabs}
 \usepackage{subcaption}
 \usepackage{tipa}
 \usepackage{pdflscape}
\usepackage{multirow}
\usepackage[T1]{fontenc}

\usepackage[utf8]{inputenc}

\usepackage{microtype}

\usepackage{inconsolata}

\usepackage{graphicx}
 \extrafloats{100}

\title{Towards a Diagnostic and Predictive Evaluation Methodology for Sequence Labeling Tasks}

\name{Elena Álvarez-Mellado, Julio Gonzalo} 

\address{NLP\&IR research group, UNED \\
         \{elena.alvarez, julio\}@lsi.uned.es\\}

\abstract{
Standard evaluation in NLP typically indicates that system A is better on average than system B, but it provides little info on how to improve performance and, what is worse, it should not come as a surprise if B ends up being better than A on outside data. 
We propose an evaluation methodology for sequence labeling tasks grounded in error analysis that provides both quantitative and qualitative information on where systems must be improved and predicts how models will perform on a different distribution.
The key is to create test sets that, contrary to common practice, do not rely on gathering large amounts of real-world in-distribution scraped data, but consists in handcrafting a small set of linguistically motivated examples that exhaustively cover the range of span attributes (such as shape, length, casing, sentence position, etc.) a system may encounter in the wild.
We demonstrate this methodology on a benchmark for anglicism identification in Spanish. 
Our methodology provides results that are \emph{diagnostic} (because they help identify systematic weaknesses in performance), \emph{actionable} (because they can inform which model is better suited for a given scenario) and \emph{predictive}: our method predicts model performance on external datasets with a median correlation of 0.85.
 \\ \newline  }

\begin{document}

\maketitleabstract

\pdfoutput=1

\section{Introduction}
Evaluation in NLP is based on comparing the aggregated scores systems obtain on a given test set for a given task. 
NLP evaluation, however, is plagued by a lingering issue: the fact that the score obtained by a system on a given test set does not tell us much about how the system can be improved, nor does it anticipate how the system will perform when tested on a different sample. 
Consequently, an excellent score on a given test set does not guarantee that the system will generalize well to a different unseen distribution \citep{poblete2019sigir}.
As a result, recent work has pointed out the need to dive deeper into the results produced by NLP models in order to get a better understanding of what models are capable and not capable of, and anticipate how models generalize to new data \citep{zhou_predictable_2023} and perform on real-world scenarios \citep{10.1162/COLI.a.18}. 

In this paper we propose a new methodology  grounded in error analysis to create test sets for sequence labeling tasks.
We apply it for the task of retrieving anglicisms from Spanish text and produce BLAS, a Benchmark for Loanwords and Anglicisms in Spanish. 
As a test set, BLAS is rather atypical. Instead of being a large collection of real-world in-distribution scraped data, BLAS is a small set of linguistically-motivated examples made from scratch and designed to exhaustively cover the range of span variability a model may encounter in the wild (in terms of shape, length, sentence position, casing, etc.) and systematically assess the different types of errors a model can make.
Our methodology produces results that are diagnostic, actionable and predictive: scores on BLAS not only serve as an evaluation benchmark to evaluate competing systems, but can also assist researchers identify which phenomena models are struggling with (thus diagnostic), which model is the best choice for a given scenario (thus actionable) and how the model will perform on a different distribution (thus predictive).

\section{The benchmark paradigm and its discontents}
In this section, we provide an overview of the current landscape of benchmark-centric evaluation in NLP and discuss several of its pitfalls.
\label{sec:discontents} 
\subsection{The benchmark paradigm}
Dataset creation is a crucial aspect of NLP research:
the availability of annotated data guides the progress in the field and the way datasets are constructed even shapes the way we frame and approach tasks \citep{paullada_data_2021}.

Various factors may influence the dataset creation process (language, annotation resources, etc).
More frequently than not, the availability of the data itself (the existence of previously compiled corpora or scrapeable texts, etc.) is a major factor in dataset creation and determines its content and the linguistic phenomena that gets represented in it \citep{plank2016non} and, in consequence, the  type of phenomena that models will be evaluated on. 

When a dataset for a given task is released, systems begin to be trained and evaluated on it. 
Certain datasets become the standard test set against which models are evaluated, thus becoming a benchmark for the task, such as CoNLL03 \citeplanguageresource{tjong-kim-sang-de-meulder-2003-introduction} or OntoNotes \citeplanguageresource{pradhan-etal-2013-towards} for NER.
Making progress on that given task now implies obtaining better scores on a particular test set.
In other words, solving a general task may now be restricted to making systems capable of modeling the phenomena that appear in one specific test set.
This means that the  phenomena that our modeling efforts will consider will be heavily influenced by the linguistic attributes that our text sample (the dataset) contained to begin with.

After reaching a certain degree of progress, improvement over scores may become elusive: more work may be required to produce increasingly smaller improvements.
Scores can eventually plateau, thus producing a glass ceiling situation \citep{stanislawek-etal-2019-named}.
Researchers and practitioners try to grasp what there is left to improve for that given task and look at the errors that their systems are making on the test set in order to identify the type of examples their models are failing at \citeplanguageresource{rueda-etal-2024-conll}.
With the development of new techniques, significant improvement on the scores may finally be reached. 
If the scores are high enough so that there is little room for improvement, the dataset can be considered saturated and sometimes the task will even be claimed to be solved \citeplanguageresource{kiela-etal-2021-dynabench}.

Eventually, a mismatch between the optimistic scores obtained by models in lab conditions over benchmark data and the results obtained by those same models on real-world scenarios will be identified \citep{lin-etal-2020-rigorous,10.1162/COLI.a.18}.
As a result, patches will be explored to account for the type of naturalistic data that models are struggling with \citep{heigold-etal-2018-robust,ribeiro-etal-2018-semantically,namysl-etal-2021-empirical}. Finally, a new and more challenging dataset will be released for the task, and the process starts again.

\subsection{Its discontents}
The Sisyphean approach we have just described, although standard, raises the following issues.

\paragraph{The limitations of identically distributed evaluation.}
Under this paradigm models are trained and tested over different sets that, although assumed to be independent, are similarly sampled and identically
distributed \citep{hupkes2023taxonomy}.
Evaluation over an identically-distributed sample may be suitable in certain restricted or predictable scenarios, but it has been shown that its results can overestimate performance and  lead to poor generalization \citep{gorman-bedrick-2019-need}.

Moreover, calling a task ``solved'' because outstanding scores were obtained on a split of a dataset where most of the data is alike and where the presence of worst-case scenarios or linguistic variations of the phenomenon in question is not ensured may be too optimistic.
Unsurprisingly, minor perturbations that pose no challenge for humans (e.g., casing or out-of-domain topics) can drastically degrade performance on allegedly solved tasks \citep{gururangan-etal-2018-annotation,mayhew-etal-2019-ner}, highlighting the lack of robustness of these models \citep{ma-etal-2023-towards}.

\paragraph{The limits of adversarial test sets.}
As a result, some voices have advocated for building adversarial test sets, that is, populating test sets with the most challenging cases possible \citep{sogaard-etal-2021-need}.
As useful as they may be, adversarial test sets are not a silver bullet. It is indeed important to know that a certain model succeeds on the worst case scenarios.
But evaluating exclusively on these types of extreme cases will prevent us from actually learning what a model can do: if tested on challenging cases that fool most models, a model that is good but far from perfect may perform as poorly as a dummy baseline.
An evaluation that aims to be useful and actionable should be informative about the models' capabilities, as well as granular enough that it can help us distinguish between a bad model, a mediocre model, a good model and a great model, something that exclusively evaluating on the most difficult cases cannot guarantee \citep{rodriguez-etal-2021-evaluation}.

\paragraph{The curse of holistic scores.}
Even if the test set is balanced in terms of the difficulty of the examples it contains, model comparison based on a score over the whole test set may be short-sighted too, because by doing so we are losing crucial information: the type of data that the model is succeeding at. 
Let's say that we have a task for which state-of-the-art models achieve a recall of 80\%.
A model that only scores 30\% will be deemed mediocre and probably irrelevant. 
But what if the data that the apparently mediocre model is succeeding at includes precisely the 20\% of data that SOTA models are failing at? 
That would be a remarkable achievement that could illuminate the areas of research future work should concentrate on, but we have no way of knowing that because in score-based evaluation is the overall score that counts, not the section of data it was achieved on.

\paragraph{Beyond error analysis: the need for success analysis.}
In order to better contextualize the information that is portrayed by numeric scores only, qualitative analysis are usually performed, generally through error analysis.
Error analysis usually consists in identifying the type of examples where models are failing and consequently establishing areas that require further work.
Although it may seem counterintuitive, focusing exclusively on errors as a way of deciding what areas of modeling require improvement may not be the best solution either: 
error analysis can only be a productive approach towards the improvement of models if it is accompanied by a \emph{success analysis} that contextualizes how prevalent certain phenomena are \citep{wu-etal-2019-errudite}.
Otherwise we risk the possibility of overstating the importance of some errors and minimizing others, which may lead to misguided decisions.

In a nutshell, it is a wide-known (yet unsolved) fact in NLP that standard evaluation based on providing a holistic score averaged over a test set (that, more often than not, will usually be identically distributed to the data contained in the training set) does not tell us much about what the model can and cannot do, which areas of the problem require further work, or how the model will generalize to a different distribution.

\section{Our methodology: scope and overview}
\label{sec:rationale}
Inspired by our observations from Section~\ref{sec:discontents}, we propose a methodology for creating test sets in which examples are selected based on their linguistic properties.
This way of proceeding radically diverges from the usual data-centric approach, in which the readiness of data (and not its linguistic characteristics) guides the data selection process. 

The scope of our methodology is restricted to span identification tasks.  
Span identification tasks are a type of sequence labeling tasks in which relevant spans are retrieved from text. 
Named entity recognition (NER) or multiword expression detection (MWE) are prime examples of span identification tasks. 
In span identification tasks, certain formal attributes of the span (such as its length or its distinctiveness) or of the context it appears in will make the task of retrieving the span more or less challenging \citep{papay-etal-2020-dissecting}.
However, none of these issues are taken into account in the standard evaluation of sequence labeling models.

We propose to populate sequence labeling test sets with examples that are selected based on the linguistic attributes of the spans they contain (in terms of shape, length, casing, position within the sentence, etc.), so that scores can be computed over examples that share the same attributes, thus producing a result that is ascribable to concrete linguistic attributes. 



The final aim of our methodology is that, whenever we see a sequence labeling model failing at retrieving a given span, we can answer the following questions: Would that model have succeeded if the same span had appeared in a more obvious context? Would it have been capable of retrieving a more prototypical span from the same context instead?
More generally: 
Which types of spans is our model finding easier to detect and which ones is it consistently failing to retrieve?
Which contexts or formal characteristics are serving as cue to the model?
How robust is our model when the cue is not present?



\section{Implementing the methodology}
\label{sec:blas}

We will demonstrate our methodology by creating  BLAS, a sequence labeling test-only dataset for anglicism identification in Spanish.




\subsection{Anglicism detection}
Anglicism detection is the task of retrieving English lexical borrowings (or \emph{anglicisms}) from non-English texts.
Anglicisms can be single-item (\textit{app}) or multiword (\textit{machine learning}, \textit{fake news}). 
The task of automatically retrieving lexical borrowings from text has proven to be useful both for lexicographic purposes and for NLP downstream tasks in various languages \citep{furiassi_retrieval_2007,alex-2008-comparing,andersen_semi-automatic_2012,losnegaard_data-driven_2012,tsvetkov-etal-2015-constraint,serigos_applying_2017} and has previously been framed as a sequence labeling task \citep{alvarez-mellado-lignos-2022-detecting,chiruzzo_overview_2023}, in which relevant in-context spans of text are retrieved from sentences (in a similar fashion to how NER or MWE are approached).

Just like in NER or MWE, in anglicism detection the retrievability of the span is heavily influenced by its shape and the context it appears in. Span attributes such as length, ambiguity, casing, surrounding context or sentence position can make the task of retrieving a given span easier or harder.
These attributes, however, are not equally represented in available datasets, as most spans tend to share the same type of formal characteristics \citep{mellado2024characterizing}.
Our aim with BLAS is to produce a test set where different combinations of span types and contexts are well represented and the characteristics that can lead to a model missing a span are exhaustively explored. 

\subsection{Span attribute selection}
\label{sec:selection}
The first step of our methodology for the data selection process is to define which span attributes should models be evaluated on (and thus be represented in our dataset). 
For BLAS, we decided to focus on the following attributes, which previous work on error analysis had identified as relevant for anglicism identification systems \citep{mellado2024characterizing}:
\begin{itemize}
    \item[-] Span length: being single-item or multiword.
    \item[-] Span shape: complying with the graphotactic expectation of the recipient language or not\footnote{Whether the anglicism violates the spelling rules of the recipient language. For instance, the anglicism \textit{streaming} does not comply with the graphotactical expectations that Spanish speakers have (because words in Spanish are not supposed to begin with \textit{str-} or end in \textit{-ing}), but \textit{online} does.}.
    \item[-] Sentence position: appearing mid sentence or in sentence initial position.
    \item[-] Quotations: being surrounded by quotation marks or not.\footnote{According to orthographical rules of Spanish, lexical borrowings such as anglicisms should be written between quotation marks or in italics.}
    \item[-] Casing: being lowercase, titlecase, etc.
    \item[-] Adjacency: being collocated to another span.
    \item[-] Ambiguity: the same sequence can be seen labeled as a span or not, depending on the context.\footnote{For example, the word \textit{pie} will be an anglicism when talking about dessert (as in \textit{un pie de limón}) but it will be a native Spanish word when it means ``foot''.}
\end{itemize}
These will be the linguistic attributes that our test set will evaluate.

\begin{table*}[t]

\tiny\centering%
\begin{tabularx}{\linewidth}{X X X r l}
\toprule
Type & Length & Position & \# sentences & Example\\
\midrule
Compliant & Single-word & Initial & 10 & \textit{\underline{Burpees} para perder kilos sin salir de casa.}\\
\cmidrule(lr){3-5}
 & & Mid & 10 & \textit{Los \underline{burpees} son efectivos para perder peso.}
 \\
\cmidrule(lr){2-5}
 & Multiword & Initial & 10  & \textit{\underline{Medal race} entre Nigeria y Polonia por la plata.}\\ 
\cmidrule(lr){3-5}
 & & Mid & 10 & \textit{Ambos países han llegado a la  \underline{medal race} ajustados de puntos.}\\
\midrule
Non-compliant & Single-word & Initial & 10 & \textit{\underline{Spoilers} del episodio siete a continuación.}\\
\cmidrule(lr){3-5}
 & & Mid & 10 & \textit{Las redes amanecieron con mensajes llenos de \underline{spoilers} del último capítulo.}\\
\cmidrule(lr){2-5}
 & Multiword & Initial & 10 & \textit{\underline{Fact checkers} confirman que la cifra aportada por el ministerio no es correcta.}\\
\cmidrule(lr){3-5}
 & & Mid & 10 & \textit{Los \underline{fact checkers} contrastarán los datos durante el debate.} \\ 
\midrule
Mixed compliant & Multiword & Initial & 10 & \textit{\underline{Joint ventures} de todo el mundo se reúnen en la mayor feria mundial de la industria.} \\
\cmidrule(lr){3-5}
 & & Mid & 10 & \textit{La fusión de ambas compañías supone un hito en la historia de las \underline{joint ventures.}}\\
 \midrule
 Ambiguous & Single-word & Mid & 3  & \textit{Receta de \underline{pie} de limón.}\\
\cmidrule(lr){2-5}
 & Multiword &  Mid & 10  & \textit{La reina escogió un conjunto \underline{total red} para el evento.}\\
\midrule
Mixed ambiguous & Multiword & Initial & 10 & \textit{\underline{Casual looks} con bufanda para esta temporada.} \\
\cmidrule(lr){3-5}
 & & Mid & 10 & \textit{Los vestidos dejan hueco a \underline{casual looks} más alegres y desenfadados.}\\
\midrule
Adjacent & Single-word & Mid & 10  & \textit{La agencia se especializa en campañas de \underline{marketing} \underline{online}.}\\ \cmidrule(lr){2-5}
 & Multiword & Mid & 10 & \textit{Ahora trabaja como \underline{head hunter} \underline{full time}.}\\
\bottomrule
\end{tabularx}
\caption{Number of sentences in BLAS per span type.}
\label{tab:sentence_num}
\end{table*}

\begin{table*}[t]
\centering
\resizebox{\textwidth}{!}{%
\begin{tabular}{llp{10cm}}
Quotation marks
   &
Casing 
   &
Example \\
  \midrule
No &  Standard &  \textit{Los \underline{burpees} son efectivos para perder peso.} \\
No &  All words are lowercase &  \textit{los \underline{burpees} son efectivos para perder peso.}\\
No &  All words are uppercase &  \textit{LOS \underline{BURPEES} SON EFECTIVOS PARA PERDER PESO.}\\
No &  All words are capitalized &  \textit{Los \underline{Burpees} Son Efectivos Para Perder Peso.}\\
No &  Borrowings are uppercase &  \textit{Los \underline{BURPEES} son efectivos para perder peso.}\\
No &  Borrowings are capitalized &  \textit{Los \underline{Burpees} son efectivos para perder peso.}\\

  \midrule
Yes &  Standard &  \textit{Los ``\underline{burpees}'' son efectivos para perder peso.} \\
Yes & All words are lowercase &  \textit{los ``\underline{burpees}'' son efectivo para perder peso.}\\
Yes &  All words are uppercase &  \textit{LOS ``\underline{BURPEES}'' SON EFECTIVOS PARA PERDER PESO.}\\
Yes & All words are capitalized &  \textit{Los ``\underline{Burpees}'' Son Efectivos Para Perder Peso.}\\
Yes &  Borrowings are uppercase &  \textit{Los ``\underline{BURPEES}'' son efectivos para perder peso.}\\
Yes &  Borrowings are capitalized &  \textit{Los ``\underline{Burpees}'' son efectivos para perder peso.}\\
\bottomrule
  \end{tabular}}
\caption{Casing and quotation mark perturbations considered in BLAS.}
\label{tab:transformations}
\end{table*}

\subsection{Data creation}
\label{sec:blas_construction}
We populated BLAS with sentences that exhaustively cover the different combinations of the span attributes we listed in Section \ref{sec:selection} (for example, non-ambiguous multiword, graphotactically-compliant uppercase spans that appear in sentence initial position between quotation marks, etc.).
Finding examples from real sources that covered all combination of attributes would have been an unmanageable endeavor, as anglicisms are a sparse phenomenon that amount to 1\%-2\% of the language \citep{gorlach_felix}.

Consequently, instead of populating our benchmark with real-world examples, we decided to have such examples written from scratch by a linguist (see Appendix \ref{sec:appendix_criteria} for guidelines on the writing process).
This approach, although unconventional, offered several benefits. 
First, instead of being confined to evaluating systems on the data that is available on the web, we gained full control of the linguistic attributes that our dataset would cover. 
Creating our examples from scratch also avoided copyright issues that usually prevent datasets from being freely reshared.
Finally, the fact that the sentences were new and not gathered from any other existing sources ensured that no model had previously been exposed to the data contained in the dataset, thus avoiding data contamination issues \citep{sainz-etal-2023-nlp}.


The mechanics behind the creation of BLAS consistently followed the same procedure for all examples: (1) identifying the span attributes we wanted to evaluate (being graphotactically compliant, being ambiguous, being multiword, etc.) (see Section~\ref{sec:selection}); (2) curating a minimal but varied list of ten spans that satisfied those attributes (see Appendix~\ref{appendix:selection}); (3) writing sentence examples for each of the selected spans in different sentence positions (see Table \ref{tab:sentence_num} and Appendix \ref{appendix:writing}); and (4) the original sentences underwent a series of systematic casing and punctuation perturbations (all text goes into lowercase, all quotation marks are removed, etc.) (see Table \ref{tab:transformations} and Appendix \ref{appendix:transformations}). 
The purpose of these perturbations is to explore all possible punctuation configurations in which a sentence and a span can appear and to control, on top of the span attribute selection, for models' overreliance on certain orthotypographic cues such as casing and quotation mark presence, which have been shown to affect model performance in sequence labeling tasks \cite{mayhew-etal-2019-ner,mellado2024characterizing}.

\subsection{Benchmark description}
The result of the process described in Section \ref{sec:blas_construction} is BLAS (Benchmark of Loanwords and Anglicisms in Spanish), a collection of 1,836 annotated sentences in Spanish (37,344 tokens) that contains 2,076 spans labeled as anglicisms. 
Every sentence in BLAS contains one span labeled as anglicism, except for the sentences with adjacent spans, which contain two.

Each sentence in BLAS is characterized by the linguistic attributes of the span it contains (in terms of shape, the context it appears in, its position within the sentence, its casing, etc.).
When models are evaluated on these sentences, scores can be computed over subsets of the data that share the same span attributes, which enables ascribing the resulting score to concrete linguistic attributes. 

In order to assess its validity, our dataset was provided to a second annotator for double annotation. 
The Cohen’s kappa IAA at token level was 0.98 (0.99 of pairwise F1 score at span level), which indicates high reliability \citep{artstein-poesio-2008-survey}.

\section{Experiments \& results}
\label{sec:blas_experiments}

\subsection{Experiments across models}
\label{sec:blas_results}
The BLAS test set that resulted from Section~\ref{sec:blas} was used to evaluate six baseline models that had previously been used for anglicism retrieval \citep{alvarez2025lexical}: five already-available supervised models that had been trained on the COALAS dataset from \citet{alvarez-mellado-lignos-2022-detecting} (a CRF, Transformer-based BETO and mBERT, and two BiLSTM-CRFs with different types of word and subword embeddings), and one additional large language model on a few-shot approach (8B-Llama3, see Appendix \ref{appendix:llama3}). As BLAS is a test-only dataset, the models were not trained on any split of BLAS, but only evaluated on it.

As all sentences in BLAS are positive examples (they all contain at least one span) and anglicisms are a sparse phenomenon, the appropriate diagnostic metric to analyze results is recall. 
As we will see in Section \ref{sec:blas_experiments}, recall will also serve as a reliable proxy to F1 score performance.



\subsubsection{Overall performance}
\label{sec:blas_overall_results}
Table \ref{tab:overall_recall_blas} displays overall recall scores obtained by the six models on BLAS. 
Scores range between 6\% and 36\%.
These scores were obtained over the whole collection of 1,836 sentences in BLAS, which include all span types with all punctuation transformations.

These results are far below the scores reported by \citet{alvarez-mellado-lignos-2022-detecting} on COALAS test set (the previously available dataset for the task), where the best performing model obtained an average recall of 78.
This gap in recall showcases that models that learned in the supervised fashion from the biases and characteristics of the training dataset struggle to generalize to outside data that is substantially different in terms of linguistic characteristics (even when the language and genre of the text is the same).
This shows that, although small in quantity, a careful selection of examples such as the one we performed in BLAS can serve to evaluate aspects and identify weaknesses in performance that go unnoticed when evaluating on large amounts of in-distribution data.

On the other side, the best-performing model on BLAS is Llama3 (R=36.37), well above all the supervised models. 
This is remarkable, as Llama3 ranked last when tested on COALAS, even below the CRF.  These results suggest that LLMs on few-shot prompting are more robust and better equipped than traditional supervised models when dealing with unseen data that is substantially different to the training data available.
If we take BLAS as an exhaustive collection of examples for evaluating the ability of models to retrieve anglicisms from a variety of linguistic shapes and contexts, then we can conclude that anglicism retrieval is far from being a solved task in Spanish, at least for the models we have explored.

\begin{table}[t]
\centering
\resizebox{\columnwidth}{!}{%
\begin{tabular}{@{}lrr|rr@{}}
\toprule
\textbf{}        &  \multicolumn{2}{c}{\textbf{BLAS}}  & \multicolumn{2}{c}{\textbf{COALAS} (\citeyear{alvarez-mellado-lignos-2022-detecting})}  \\ \midrule
\textbf{CRF} &  6.50 & \#6  & 43.04  & \#5 \\
\textbf{BETO}  &  23.55 & \#4  &  75.50 & \#4 \\
\textbf{mBERT} &    23.80 & \#2 & 76.16  & \#3 \\
\textbf{BiLSTM-CRF (unadapted)} & 23.55 & \#4 &  78.34  & \#2 \\
\textbf{BiLSTM-CRF (codeswitch)}    &  23.75 & \#3 &  \bf 78.72 & \#1  \\
\textbf{Llama3} & \bf  36.37 & \#1 &  32.14 & \#6  \\ 
\bottomrule
\end{tabular}%
}
\caption{Span recall per model on BLAS and COALAS.}
\label{tab:overall_recall_blas}
\end{table}

\begin{table*}[t]
\centering
\resizebox{\textwidth}{!}{%
\begin{tabular}{@{}l|ll|ll|ll|ll|ll|ll|ll|@{}}
\cmidrule(l){2-13}

                                                                                            & \multicolumn{2}{c|}{\textbf{CRF}}                           & \multicolumn{2}{c|}{\textbf{BETO}}                                   & \multicolumn{2}{c|}{\textbf{mBERT}}                              & \multicolumn{2}{c|}{\textbf{BiLSTM (unadp)}}                         & \multicolumn{2}{c|}{\textbf{BiLSTM (codeswitch)}}                 & \multicolumn{2}{c|}{\textbf{Llama3}}\\ 
                                                                                      \cmidrule(l){2-13}      
\multicolumn{1}{l|}{}                                                                       & With quot.                & \multicolumn{1}{l|}{W/o quot.} & With quot.                         & \multicolumn{1}{l|}{W/o quot.} & With quot.                & \multicolumn{1}{l|}{W/o quot.}      & With quot.                         & \multicolumn{1}{l|}{W/o quot.} & With quot.                & \multicolumn{1}{l|}{W/o quot.}      & With quot.                         & \multicolumn{1}{l|}{W/o quot.}   \\ \midrule
\multicolumn{1}{l|}{\textbf{Standard casing}}                                               & \multicolumn{1}{r}{21.97} & \multicolumn{1}{r|}{7.51}      & \multicolumn{1}{r}{\textbf{68.79}} & \multicolumn{1}{r|}{45.09}     & \multicolumn{1}{r}{61.27} & \multicolumn{1}{r|}{\textbf{49.13}} & \multicolumn{1}{r}{66.47}          & \multicolumn{1}{r|}{45.09}     & \multicolumn{1}{r}{60.12} & \multicolumn{1}{r|}{46.24}          & \multicolumn{1}{r}{66.47}          & \multicolumn{1}{r|}{23.12}        \\
\multicolumn{1}{l|}{\textbf{\begin{tabular}[c]{@{}l@{}}Text is   lowercase\end{tabular}}} & \multicolumn{1}{r}{31.79} & \multicolumn{1}{r|}{16.18}     & \multicolumn{1}{r}{87.86}          & \multicolumn{1}{r|}{56.65}     & \multicolumn{1}{r}{83.82} & \multicolumn{1}{r|}{65.32}          & \multicolumn{1}{r}{\textbf{90.17}} & \multicolumn{1}{r|}{61.27}     & \multicolumn{1}{r}{80.92} & \multicolumn{1}{r|}{\textbf{68.21}} & \multicolumn{1}{r}{67.63}          & \multicolumn{1}{r|}{27.75}  \\
\multicolumn{1}{l|}{\textbf{\begin{tabular}[c]{@{}l@{}}Span is   titlecase\end{tabular}}} & \multicolumn{1}{r}{0.00}  & \multicolumn{1}{r|}{0.00}      & \multicolumn{1}{r}{5.20}           & \multicolumn{1}{r|}{5.20}      & \multicolumn{1}{r}{4.62}  & \multicolumn{1}{r|}{5.78}           & \multicolumn{1}{r}{6.36}           & \multicolumn{1}{r|}{5.78}      & \multicolumn{1}{r}{5.78}  & \multicolumn{1}{r|}{6.36}           & \multicolumn{1}{r}{\textbf{56.65}} & \multicolumn{1}{r|}{\textbf{24.86}}\\
\multicolumn{1}{l|}{\textbf{\begin{tabular}[c]{@{}l@{}}Text is   titlecase\end{tabular}}} & \multicolumn{1}{r}{0.00}  & \multicolumn{1}{r|}{0.00}      & \multicolumn{1}{r}{7.51}           & \multicolumn{1}{r|}{5.78}      & \multicolumn{1}{r}{10.40} & \multicolumn{1}{r|}{5.20}           & \multicolumn{1}{r}{4.05}           & \multicolumn{1}{r|}{2.89}      & \multicolumn{1}{r}{6.94}  & \multicolumn{1}{r|}{6.94}           & \multicolumn{1}{r}{\textbf{32.95}} & \multicolumn{1}{r|}{\textbf{9.25}} \\
\multicolumn{1}{l|}{\textbf{\begin{tabular}[c]{@{}l@{}}Span is   uppercase\end{tabular}}} & \multicolumn{1}{r}{0.00}  & \multicolumn{1}{r|}{0.58}      & \multicolumn{1}{r}{0.00}           & \multicolumn{1}{r|}{0.58}      & \multicolumn{1}{r}{0.00}  & \multicolumn{1}{r|}{0.00}           & \multicolumn{1}{r}{0.00}           & \multicolumn{1}{r|}{0.00}      & \multicolumn{1}{r}{1.16}  & \multicolumn{1}{r|}{2.31}           & \multicolumn{1}{r}{\textbf{54.91}} & \multicolumn{1}{r|}{\textbf{45.09}} \\
\multicolumn{1}{l|}{\textbf{\begin{tabular}[c]{@{}l@{}}Text is  uppercase\end{tabular}}} & \multicolumn{1}{r}{0.00}  & \multicolumn{1}{r|}{0.00}      & \multicolumn{1}{r}{0.00}           & \multicolumn{1}{r|}{0.00}      & \multicolumn{1}{r}{0.00}  & \multicolumn{1}{r|}{0.00}           & \multicolumn{1}{r}{0.00}           & \multicolumn{1}{r|}{0.58}      & \multicolumn{1}{r}{0.00}  & \multicolumn{1}{r|}{0.00}           & \multicolumn{1}{r}{\textbf{21.97}} & \multicolumn{1}{r|}{\textbf{5.78}}  \\ \bottomrule
\end{tabular}%
}
\caption{Overall span recall on BLAS per model on different casing and quotation mark configurations.}
\label{tab:overall_recall_configurations}
\end{table*}

\begin{table*}[t]
\centering
\resizebox{\textwidth}{!}{%
\begin{tabular}{@{}lll|rr|rr|rr|rr|rr|rr|@{}}
\cmidrule(l){4-15}
                                 &                         &                   & \multicolumn{2}{c|}{\textbf{CRF}}                               & \multicolumn{2}{c|}{\textbf{BETO}}                              & \multicolumn{2}{c|}{\textbf{mBERT}}                             & \multicolumn{2}{c|}{\textbf{BiLSTM (unad)}}                     & \multicolumn{2}{c|}{\textbf{BiLSTM (codeswitch)}}               & \multicolumn{2}{c|}{\textbf{Llama3}}                            \\ \toprule
\textbf{Type}                    & \textbf{Length}         & \textbf{Position} & \multicolumn{1}{l}{With quot.} & \multicolumn{1}{l|}{W/o quot.} & \multicolumn{1}{l}{With quot.} & \multicolumn{1}{l|}{W/o quot.} & \multicolumn{1}{l}{With quot.} & \multicolumn{1}{l|}{W/o quot.} & \multicolumn{1}{l}{With quot.} & \multicolumn{1}{l|}{W/o quot.} & \multicolumn{1}{l}{With quot.} & \multicolumn{1}{l|}{W/o quot.} & \multicolumn{1}{l}{With quot.} & \multicolumn{1}{l|}{W/o quot.} \\ \midrule
\multirow{4}{*}{compliant}       & \multirow{2}{*}{multi}  & ini               & 0.00                           & 0.00                           & 40.00                          & \textbf{50.00}                          & 20.00                          & 40.00                          & 20.00                          & 40.00                          & 30.00                          & 30.00                          & \textbf{90.00}                          & 20.00                          \\
                                 &                         & mid               & 20.00                          & 0.00                           & \textbf{100.00}                         & 80.00                          & 90.00                          & 80.00                          & \textbf{100.00}                         & 80.00                          & 90.00                          & \textbf{90.00}                          & 80.00                          & 30.00                          \\
                                 & \multirow{2}{*}{single} & ini               & 0.00                           & 0.00                           & 10.00                          & \textbf{20.00}                          & 0.00                           & 10.00                          & 20.00                          & \textbf{20.00}                          & 10.00                          & \textbf{20.00}                          & \textbf{50.00}                          & 10.00                          \\
                                 &                         & mid               & 10.00                          & 0.00                           & 70.00                          & 50.00                          & 80.00                          & 80.00                          & 90.00                          & 80.00                          & \textbf{100.00}                         & \textbf{90.00}                          & 40.00                          & 10.00                          \\ \midrule
\multirow{4}{*}{non compliant}   & \multirow{2}{*}{multi}  & ini               & 10.00                          & 10.00                          & \textbf{90.00}                          & 50.00                          & 60.00                          & \textbf{70.00}                          & 30.00                          & 40.00                          & 20.00                          &\textbf{ 70.00}                          & \textbf{90.00}                          & 30.00                          \\
                                 &                         & mid               & 60.00                          & 10.00                          & \textbf{100.00}                         & 80.00                          &\textbf{ 100.00 }                        & \textbf{100.00 }                        & \textbf{100.00 }                        & \textbf{100.00}                         & \textbf{100.00}                         & \textbf{100.00}                         & 90.00                          & 40.00                          \\
                                 & \multirow{2}{*}{single} & ini               & 0.00                           & 0.00                           & 20.00                          & 20.00                          & 20.00                          & 30.00                          & 20.00                          & 10.00                          & 10.00                          & \textbf{40.00}                          &\textbf{ 80.00}                          & \textbf{40.00}                          \\
                                 &                         & mid               & 50.00                          & 40.00                          &\textbf{ 100.00}                         & \textbf{100.00}                         & \textbf{100.00}                         & 90.00                          & \textbf{100.00}                         & 90.00                          & \textbf{100.00}                         & \textbf{100.00}                         & 60.00                          & 50.00                          \\ \midrule
\multirow{2}{*}{mixed compliant} & \multirow{2}{*}{multi}  & ini               & 0.00                           & 0.00                           & 30.00                          & \textbf{40.00}                          & 20.00                          & 30.00                          & 20.00                          & 30.00                          & 20.00                          & 20.00                          &\textbf{ 80.00 }                         & 20.00                          \\
                                 &                         & mid               & 10.00                          & 0.00                           & \textbf{100.00}                         & 90.00                          & 90.00                          & 90.00                          & \textbf{100.00 }                        & \textbf{100.00}                         & 90.00                          & 80.00                          & 70.00                          & 30.00                          \\ \midrule
\multirow{2}{*}{ambiguous}       & multi                   & mid               & 0.00                           & 0.00                           & 50.00                          & 30.00                          & 50.00                          & \textbf{40.00}                          &\textbf{ 70.00}                          & 20.00                          & 50.00                          & 10.00                          & \textbf{70.00}                          & 10.00                          \\
                                 & single                  & mid               & 0.00                           & 0.00                           &\textbf{ 33.33}                          &\textbf{ 33.33}                          & \textbf{33.33}                          & 0.00                           & \textbf{33.33  }                        & 0.00                           & 0.00                           & 0.00                           & \textbf{33.33}                          & 0.00                           \\ \midrule
\multirow{2}{*}{mixed ambiguous} & \multirow{2}{*}{multi}  & ini               & 0.00                           & 0.00                           & 30.00                          & \textbf{40.00}                          & 30.00                          & 30.00                          & 20.00                          & 20.00                          & 0.00                           & 0.00                           & \textbf{90.00}                          & 20.00                          \\
                                 &                         & mid               & 10.00                          & 0.00                           & 70.00                          & 50.00                          & 70.00                          & 50.00                          & \textbf{100.00}                         & \textbf{60.00 }                         & 60.00                          & 50.00                          & 80.00                          & 40.00                          \\ \midrule
\multirow{2}{*}{adjacent}        & multi                   & mid               & 45.00                          & 10.00                          & \textbf{90.00 }                         & 20.00                          & 65.00                          & \textbf{30.00   }                       & 80.00                          & 5.00                           & 80.00                          & 15.00                          & 55.00                          & 10.00                          \\
                                 & single                  & mid               & 60.00                          & 25.00                          & 95.00                          & 15.00                          & 95.00                          & 25.00                          & 95.00                          & \textbf{40.00}                          &\textbf{ 100.00  }                       & 35.00                          & 30.00                          & 15.00                          \\ \bottomrule
\end{tabular}%
}
\caption{Span recall obtained by each model over BLAS over different types of spans and contexts (on standard casing, with and without quotation marks).}
\label{tab:master_standard}
\end{table*}

\subsubsection{Performance across punctuation}
Holistic results such as the ones reported on Table~\ref{tab:overall_recall_blas}, however, do not tell us much about the type of examples models are failing to retrieve, or how to fix them.
Table \ref{tab:overall_recall_configurations} displays the same results than Table~\ref{tab:overall_recall_blas}, but split into the different casing and quotation mark configurations of BLAS. 

Two facts stand out from Table \ref{tab:overall_recall_configurations}. 
First, all models obtained better results on the versions of the dataset where the spans were written between quotation marks. 
The differences in recall between examples with quotation marks and without them are substantial, with scores dropping even around 20 points or more for some models. 
Our results reveal that Llama3 was by far the most vulnerable to quotation mark absence, with a loss of more than 40 points in recall when quotation marks were not present on the standard casing split. 
Note that these differences in scores are fully attributable to the absence of quotation marks, because both splits contain exactly the same sentences, but in one split the spans are surrounded by quotation marks, while in the other they are not.
In other words, the very same models that were capable of correctly identifying \emph{burpees} in \emph{Los ``burpees'' son efectivos para perder peso} failed to do so when the quotation marks were absent (\emph{Los burpees son efectivos para perder peso}). 

Second, supervised models consistently obtained  catastrophic results on the capitalized sections of BLAS, with models scoring zero across uppercase configurations.
Llama3, on the other hand, while it still saw a drop in performance over the capitalized sections of BLAS (especially when the whole text was affected), produced \emph{less} bad results than the supervised models, which explains its good position in the overall ranking from Table~\ref{tab:overall_recall_blas}.

The results from Table \ref{tab:overall_recall_configurations} provide us with actionable conclusions that can inform decision-making when choosing which system to use for anglicism detection on different scenarios: if we are expecting well-edited text with standard casing and quotation marks, then BETO would probably be the best model.
If quotation marks are not expected, then mBERT would be a better call.
This casing vulnerability points to an issue that future work should address: none of these models is reliable when applied to texts with non-standard capitalization.
However, if we had to choose one model for such scenario, Llama3 would be our safest bet. If we are expecting lowercase text (such as social media text), then the BiLSTMs are the best choice.

\subsubsection{Performance across different span types}
\label{sec:blas_results_model}
We now analyze performance across different span types.
We will focus on the standard casing configuration (where only proper names and sentence-initial words are capitalized), as it is the most canonical and results over the capitalized sections of BLAS were catastrophic across most models.


Table \ref{tab:master_standard} shows that certain span attributes produce better results than others, like mid-sentence spans. 
In fact, non-compliant spans that appear in mid-sentence position with quotation marks (such as \textit{spoilers} in \textit{Las redes amancieron con mensajes llenos de ``spoilers''  del último capítulo}) can  essentially be considered solved, with almost all models producing perfect or near perfect scores over those examples. 



On the other hand, performance drops substantially when retrieving spans from sentence initial position (such as \textit{spoilers} in \textit{``Spoilers'' del episodio siete a continuación}).
It is remarkable that the very same spans that were retrieved with a score of 80, 90 or even 100 when they appeared mid-sentence, produced a recall of 20, 10 or even 0 when they appeared at the beginning of the sentence.
Again, it should be noted these differences in recall between sentence-initial and mid-sentence spans can be fully ascribed to the differences in the contexts they appear in, as the spans were the same across both groups (see Table \ref{tab:sentence_num}). 

Ambiguous spans also produced substantially worse results than other types of spans, although Llama3 proved to be more robust to ambiguity, as long as quotation marks were present.
On the other hand, adjacency was not a challenge for supervised models: most of them produced scores over 80, at least with quotation marks. Llama3, on the contrary, obtained results on adjacent spans that were in line or even below the CRF.

\subsection{Predictive ability of BLAS}
The natural question that follows from our results from Section \ref{sec:blas_results} is: are the results per attribute obtained on BLAS representative of the ability of the models to perform on those linguistic attributes, not just over BLAS data, but in general? Could we use the scores obtained on BLAS to predict the overall recall that those same models would obtain on a different distribution? 
For instance, based on the results obtained on BLAS, we know that Llama3 produces a span recall of 90 over non-compliant multiword spans when they appear mid sentence and are written with quotation marks on standard casing.
Could we extrapolate those results obtained by Llama3 on BLAS to other datasets and claim that that Llama3 will successfully retrieve 90\% of the spans that share those same characteristics when tested on a different distribution?




\begin{table*}[]
\centering
\resizebox{\textwidth}{!}{
\begin{tabular}{@{}l|r|rrrr|rrrrrrrrrrrrrrrr@{}}
\toprule
\multicolumn{1}{l|}{\multirow{2}{*}{\textbf{Models}}} & \multicolumn{1}{l|}{\textbf{BLAS}} & \multicolumn{4}{c|}{\textbf{COALAS (test set)}}                                                                             & \multicolumn{4}{c|}{\textbf{COALAS (dev set)}}                                                                                   & \multicolumn{4}{c|}{\textbf{CALCS (test set A)}}                                                                                 & \multicolumn{4}{c|}{\textbf{CALCS (test set B)}}                                                                                 & \multicolumn{4}{c|}{\textbf{CALCS (dev set)}}                                                                                    \\ \cmidrule(l){2-22} 
\multicolumn{1}{c|}{}                                 & \multicolumn{1}{c|}{\textbf{R}}    & \multicolumn{2}{c}{\textbf{Predict R}}  & \multicolumn{2}{c|}{\textbf{True R}} & \multicolumn{2}{c}{\textbf{Predict R}}  & \multicolumn{2}{c|}{\textbf{True R}}  & \multicolumn{2}{c}{\textbf{Predict R}}  & \multicolumn{2}{c|}{\textbf{True R}}  & \multicolumn{2}{c}{\textbf{Predict R}} & \multicolumn{2}{c|}{\textbf{True R}}   & \multicolumn{2}{c}{\textbf{Predict R}}  & \multicolumn{2}{c|}{\textbf{True R}}    \\ \midrule
\textbf{CRF}                                          & 6.50                               & 26.58                                  & \#6                   & 44.31                               & \#5                    & 33.13                                  & \#6                   & 68.63                               & \multicolumn{1}{r|}{\#5}    & 29.00                                  & \#6                   & 65.57                               & \multicolumn{1}{r|}{\#5}    & 24.76                                  & \#6                   & 53.17                               & \multicolumn{1}{r|}{\#5}    & 28.99                                  & \#6                   & 64.84                               & \multicolumn{1}{r|}{\#5}    \\
\textbf{BETO}                                         & 23.55                              & 86.87                                  & \#3                   & 77.99                               & \#4                    & 87.62                                  & \#3                   & 84.05                               & \multicolumn{1}{r|}{\#4}    & 83.58                                  & \#4                   & 83.49                               & \multicolumn{1}{r|}{\#3}    & 79.13                                  & \#4                   & 81.75                               & \multicolumn{1}{r|}{\#4}    & 81.60                                  & \#4                   & 86.30                               & \multicolumn{1}{r|}{\#1}    \\
\textbf{mBERT}                                        & 23.80                              & 86.86                                  & \#4                   & 78.85                               & \#3                    & 87.10                                  & \#4                   & 84.31                               & \multicolumn{1}{r|}{\#3}    & 83.86                                  & \#3                   & 85.85                               & \multicolumn{1}{r|}{\#1}    & 83.17                                  & \#3                   & 84.13                               & \multicolumn{1}{r|}{\#3}    & 82.73                                  & \#3                   & 86.30                               & \multicolumn{1}{r|}{\#1}    \\
\textbf{BiLSTM*}                                & 23.55                              & 87.86                                  & \#2                   & 80.88                               & \#2                    & 88.73                                  & \#2                   & 89.05                               & \multicolumn{1}{r|}{\#1}    & 84.85                                  & \#2                   & 82.08                               & \multicolumn{1}{r|}{\#4}    & 84.04                                  & \#2                   & 84.92                               & \multicolumn{1}{r|}{\#2}    & 83.06                                  & \#2                   & 83.11                               & \multicolumn{1}{r|}{\#4}    \\
\textbf{BiLSTM**}                                  & 23.75                              & 92.96                                  & \#1                   & 85.76                               & \#1                    & 91.08                                  & \#1                   & 86.57                               & \multicolumn{1}{r|}{\#2}    & 89.43                                  & \#1                   & 83.96                               & \multicolumn{1}{r|}{\#2}    & 89.52                                  & \#1                   & 85.71                               & \multicolumn{1}{r|}{\#1}    & 87.53                                  & \#1                   & 85.84                               & \multicolumn{1}{r|}{\#3}    \\
\textbf{Llama3}                                       & 36.37                              & 42.93                                  & \#5                   & 33.33                               & \#6                    & 55.46                                  & \#5                   & 53.27                               & \multicolumn{1}{r|}{\#6}    & 48.53                                  & \#5                   & 53.30                               & \multicolumn{1}{r|}{\#6}    & 43.17                                  & \#5                   & 40.48                               & \multicolumn{1}{r|}{\#6}    & 48.49                                  & \#5                   & 50.68                               & \multicolumn{1}{r|}{\#6}    \\ \midrule
\textbf{Correl}                                  & \multicolumn{1}{l|}{}              & \multicolumn{1}{l}{}                   & \multicolumn{1}{l}{} & 0.94                                & 0.89                  & \multicolumn{1}{l}{}                   & \multicolumn{1}{l}{} & 0.79                                & \multicolumn{1}{r|}{0.83} & \multicolumn{1}{l}{}                   & \multicolumn{1}{l}{} & 0.85                                & \multicolumn{1}{r|}{0.66} & \multicolumn{1}{l}{}                   & \multicolumn{1}{l}{} & 0.91                                & \multicolumn{1}{r|}{0.94} & \multicolumn{1}{l}{}                   & \multicolumn{1}{l}{} & 0.84                                & \multicolumn{1}{r|}{0.41} \\ \bottomrule
\end{tabular}%
}
\caption{Correlation between the predicted and true recall scores and ranking over five datasets (Pearson and Spearman correlation coefficients). }
\label{tab:corr}
\end{table*}

In order to answer these questions, we ran the six models from Section \ref{sec:blas_results} on five external datasets: the test and dev sets from COALAS \citep{alvarez-mellado-lignos-2022-detecting} and the dev set and test sets A and B from CALCS \citep{alvarez-mellado-2020-annotated}. These are all publicly available datasets for anglicism identification in Spanish that, although similar in genre to BLAS (Spanish journalistic text), show a very different distribution in terms of the linguistic attributes represented \citep{mellado2024characterizing}. 
In addition, COALAS test set and CALCS test set B have been reported to be substantially different (in terms of out-of-distribution topics and vocabulary) to their training splits. 

The purpose of our experiment was to test if the scores the models obtained on BLAS across different span attributes could predict the results those same models would obtain on these external datasets.
To do so, we counted the number of anglicism spans per type in each of these external datasets and calculated the expected number of spans that would be successfully retrieved by simply extrapolating from the recall results obtained per type by each model on BLAS. 
Going back to our example above: there are 20 spans in the COALAS test set that are non-compliant multiword and that appear with standard casing, quotation marks and in mid sentence position. 
Given that the span recall of Llama3 obtained on BLAS on that particular type of spans is 90, we hypothesized that 90\% of those 20 spans in COALAS would be successfully retrieved by Llama3, which amount to 18 true positives and 2 false negatives.
We did the same for each type of span in all five datasets for every model, and calculated the expected overall span recall per model on each dataset based on BLAS scores (see Appendix \ref{appendix:spans_COALAS}).

Table \ref{tab:corr} compares the true recall and ranking position and the predicted recall and ranking position derived from extrapolating the results produced on BLAS. 
Results show that our method accurately predicts both recall values (with a median Pearson correlation of 0.85) and system ranking (median correlation of 0.83).
The highest correlations were obtained on COALAS test set and CALCS test set B (0.94 and 0.91, respectively), precisely the two splits that had been reported to be most dissimilar to their training splits.
Our predicted rankings over recall also display a strong correlation with F1 score rankings (median correlation 0.83).
It should be noted that overall scores on BLAS are not predictive themselves: it is the projection of the system performance across span attributes in BLAS that correlates with true scores (see Tables \ref{tab:overall_recall_blas} and \ref{tab:corr}).


\section{Discussion}
Our experiments show that BLAS can uncover systematic weaknesses in anglicism detection models that may go unnoticed when evaluating on naturalistic in-distribution data.
The fine-grained results produced on BLAS can also inform decision-making when choosing which is the best model for a given data scenario (punctuation, capitalization, types of spans, etc).
In addition to serving as a test set to evaluate models and identify systematic weaknesses, our results show that BLAS has predictive ability:
the scores per attribute obtained by six models on BLAS were able to accurately predict the performance of those models on five external test sets with a median correlation of 0.85. 
 
Although we have built our case on anglicism detection, our methodology could  in principle be applied to other span identification tasks, such as NER, where ambiguous and adjacent spans are ubiquitous and span retrievability is also sensitive to context, punctuation and sentence position.  

The graphotactic compliance distinction may be  specific to anglicism identification, but this linguistic attribute is simply a concrete instantiation of a more general issue that is common to other sequence labeling tasks: the fact that in any span-based task, there will be some spans that will be more salient  than others, a feature that has also been reported to be key in NER \citep{papay-etal-2020-dissecting,lin-etal-2021-rockner,vajjala-balasubramaniam-2022-really}.

The predictive ability displayed by BLAS fine-grained results suggests that linguistically-aided test set creation informed by error analysis may be a workable solution to the long quest for evaluation methodologies that can anticipate how a system will perform on new data, a feature that is considered key to measure the generalization abilities of NLP systems \citep{hupkes2023taxonomy,zhou_predictable_2023,10.1162/COLI.a.18}.

\section{Relation to previous work}
The ideas that guided the creation of BLAS take inspiration from previous work on robustness evaluation in NLP, especially on NER models. 

The gap in performance between the results obtained by NLP models when they are evaluated on benchmark data and when they are tested on real-world scenarios has previously been discussed \citep[e.g.][]{poblete2019sigir,kiela-etal-2021-dynabench,10.1162/COLI.a.18}.

\citet{lin-etal-2020-rigorous} referred to this gap as the difference in performance between \emph{regular NER} (NER performed in lab conditions on benchmark data) and \emph{open NER} (NER performed on real world data) and pointed out three possible reasons behind this gap: strong name regularity in NER benchmark entities (i.e., most spans have a similar shape), overlap between the entities in the test set and the training set and test instances that are rich in context.

Additional research on NER has added two more possible culprits for the lack of robustness of NER models: label inconsistency \citep{fu-etal-2020-interpretable,bernier-colborne-langlais-2020-hardeval,tu-lignos-2021-tmr} and presence of adjacent spans \citep{rueda-etal-2024-conll}.
These phenomena that plague NER performance are no different to the type of issues that we encounter in anglicism detection.
With BLAS, we have translated these issues (span regularity, ambiguity, context and adjacency) to the task of anglicism detection by selecting test examples in which these attributes are exhaustively represented. 

The idea of applying systematic perturbations to inputs in order to assess models' robustness is not new and has previously been applied to different aspects of NER performance, particularly for casing \citep{mayhew-etal-2019-ner}, span shape \citep{agarwal_entity-switched_2021,lin-etal-2021-rockner,vajjala-balasubramaniam-2022-really}, type  or length \citep{wang-etal-2021-textflint}.
It also builds on previous work on stress test evaluation of models' robustness on specific linguistic phenomena \citep{naik-etal-2018-stress,wang_superglue_2019,ribeiro-etal-2020-beyond,fu-etal-2020-interpretable}, but specifically targeting sequence labeling tasks. 
Our motivation is similar to the work of \citet{kovatchev-lease-2024-benchmark}, but they opt for describing dataset items in terms of extrinsic features (how a set of systems perform on each of the instances), while our dimensions are intrinsic and linguistically motivated: they do not depend on the set of systems tested.  

The idea of building a testbed by creating linguistically-motivated examples by hand draws from other expert-authored benchmarks from the NLU literature \citep{bowman-dahl-2021-will}, such as the FraCaS test suite \citeplanguageresource{cooper_using_1996} or the the Winograd Schema Challenge \citeplanguageresource{levesque_winograd_2012}.
It also follows the recommendations stated in \citet{linzen-2020-accelerate} of producing test-only benchmarks to avoid unintended statistical regularities, and those from \citet{sogaard-etal-2021-need} that proposes evaluating models on biased splits as evaluation protocol to determine data characteristics that affect performance.
Similarly, our methodology implements and systematizes the spirit behind other evaluation campaigns such as \citet{kiela-etal-2021-dynabench} and \citet{ettinger-etal-2017-towards}, which put the linguistic expertise at the center of NLP evaluation.

\section{Conclusions}
We introduced a new methodology grounded in error analysis to evaluate sequence labeling tasks, in which test sets are populated with examples selected based on their linguistic attributes and sampled according to the types of errors a system may make. We applied this methodology to create BLAS, a test-only dataset for anglicism identification in Spanish. Results on BLAS are more informative than scores on standard test sets, as they diagnose systematic weaknesses in performance, identify which systems work best in a given scenario, and predict how a system will generalize to new data.

\section{Limitations}
\label{sec:limitations}

\paragraph{The task}
We have demonstrated our methodology on one task only: anglicism identification in Spanish. 
Although our task is quite niche, we believe that the same methodology could be applied to other span identification tasks, such as NER and MWE, where span retrieval is affected by the very same phenomena we have explored (ambiguity, casing, sentence position, span length, span shape, etc.).
Future work should examine error analysis on these other tasks and consider whether additional attributes or modifications of the current ones are needed.

\paragraph{The type of problem}
The methodology we have presented can only be applied to span identification tasks, where spans are retrieved from in-context sentences.
It is not clear how a similar methodology could be devised for text classification problems, or what attributes should be taken into account. 

\paragraph{The metrics}
Our work has focused on evaluating the capability of models at retrieving true spans of different types and shapes from different contexts. 
Consequently, our analysis and the creation of the dataset itself focused on recall.
Prioritizing recall is suitable for tasks in which minimizing the number of false negatives is crucial (even if it is at the cost of having some false positives), such as when building technology to extract linguistic phenomena that is rare \citep{kermes-2004-text}, like lexical borrowing. 

Other tasks, however, may require taking precision into account and may wish to perform a similar evaluation over false positives. 
To evaluate precision in a similar way as we did with recall with BLAS, we would first need to investigate the type of linguistic phenomena that characterize false positive spans and then populate our test set with non-relevant spans that could pass for one.

It should be noted that predicting a precision score over an external dataset as we did with recall with BLAS would not be feasible with our methodology: in a span-based task, the sum of false negatives and true positives is known, as it is the number of total spans in the goldstandard. Therefore, we can calculate the proportion of spans that are expected to be missed by a model, based on the recall scores it obtained on the benchmark. However, if we tried to do the same with precision, even if we had a precision-based benchmark, it would be unclear what the set of potential false positive spans of the external test set would be. 

\paragraph{The authorship}
Sentences in BLAS were authored by a single linguist.
Although extensive guidelines were compiled to ensure reproducibility of our method and the double annotation yielded very high IAA, this fact may introduce biases in the data. 
Further work is required to assess for generalization if a larger multi-authored test set was to be created.

\paragraph{The dataset}
BLAS should not be taken as a definite benchmark for anglicism identification in Spanish, nor as the sole test set against a model should be measured.
Quite on the contrary, BLAS is intended as a very specific diagnosing testbed that may contribute documenting and assessing a model's capability. 
In that regard, BLAS follows the trend of other auxiliary validation sets \citep{ghaddar-etal-2021-context}.
How relevant the scores obtained by a model on BLAS are will be determined by the use case scenario where the model in question is expected to be deployed.
\section{Data availability}\label{sec:data}
BLAS was the test set used in the ADoBo\footnote{\url{https://adobo-task.github.io}} shared task on automatic detection of anglicisms at IberLEF 2025 \cite{alvarez2025overview} and the dataset is publicly  available\footnote{\url{https://github.com/lirondos/blas}}.

\section{Bibliographical References}\label{sec:reference}

\bibliographystyle{lrec2026-natbib}
\bibliography{lrec2026-example,anthology}

\begin{thebibliography}{6}
\expandafter\ifx\csname natexlab\endcsname\relax\def\natexlab#1{#1}\fi

\bibitem[{Cooper et~al.(1996)Cooper, Crouch, Van~Eijck, Fox, Jaspars, Kamp, Milward, Pinkal, Poesio, and {others}}]{cooper_using_1996}
Cooper, Robin and Crouch, Dick and Van Eijck, Jan and Fox, Chris and Jaspars, Jan and Kamp, Hans and Milward, David and Pinkal, Manfred and Poesio, Massimo and {others}. 1996.
\newblock \emph{Using the framework}.

\bibitem[{Kiela et~al.(2021)Kiela, Bartolo, Nie, Kaushik, Geiger, Wu, Vidgen, Prasad, Singh, Ringshia, Ma, Thrush, Riedel, Waseem, Stenetorp, Jia, Bansal, Potts, and Williams}]{kiela-etal-2021-dynabench}
Kiela, Douwe and Bartolo, Max and Nie, Yixin and Kaushik, Divyansh and Geiger, Atticus and Wu, Zhengxuan and Vidgen, Bertie and Prasad, Grusha and Singh, Amanpreet and Ringshia, Pratik and Ma, Zhiyi and Thrush, Tristan and Riedel, Sebastian and Waseem, Zeerak and Stenetorp, Pontus and Jia, Robin and Bansal, Mohit and Potts, Christopher and Williams, Adina. 2021.
\newblock \href {https://doi.org/10.18653/v1/2021.naacl-main.324} {\emph{Dynabench: Rethinking Benchmarking in {NLP}}}.
\newblock Association for Computational Linguistics.

\bibitem[{Levesque et~al.(2012)Levesque, Davis, and Morgenstern}]{levesque_winograd_2012}
Levesque, Hector J. and Davis, Ernest and Morgenstern, Leora. 2012.
\newblock \emph{The {Winograd} schema challenge}.
\newblock AAAI Press, {KR}'12.

\bibitem[{Pradhan et~al.(2013)Pradhan, Moschitti, Xue, Ng, Bj{\"o}rkelund, Uryupina, Zhang, and Zhong}]{pradhan-etal-2013-towards}
Pradhan, Sameer and Moschitti, Alessandro and Xue, Nianwen and Ng, Hwee Tou and Bj{\"o}rkelund, Anders and Uryupina, Olga and Zhang, Yuchen and Zhong, Zhi. 2013.
\newblock \href {https://aclanthology.org/W13-3516/} {\emph{Towards Robust Linguistic Analysis using {O}nto{N}otes}}.
\newblock Association for Computational Linguistics.

\bibitem[{Rueda et~al.(2024)Rueda, Alvarez-Mellado, and Lignos}]{rueda-etal-2024-conll}
Rueda, Andrew and Alvarez-Mellado, Elena and Lignos, Constantine. 2024.
\newblock \href {https://aclanthology.org/2024.lrec-main.330/} {\emph{{C}o{NLL}{\#}: Fine-grained Error Analysis and a Corrected Test Set for {C}o{NLL}-03 {E}nglish}}.
\newblock ELRA and ICCL.

\bibitem[{Tjong Kim~Sang and De~Meulder(2003)}]{tjong-kim-sang-de-meulder-2003-introduction}
Tjong Kim Sang, Erik F. and De Meulder, Fien. 2003.
\newblock \href {https://aclanthology.org/W03-0419/} {\emph{Introduction to the {C}o{NLL}-2003 Shared Task: Language-Independent Named Entity Recognition}}.

\end{thebibliography}

\section{Language Resource References}
\label{lr:ref}
\bibliographystylelanguageresource{lrec2026-natbib}
\bibliographylanguageresource{languageresource}

\appendix

\section{Guidelines for the creation of examples in BLAS}
\label{sec:appendix_criteria}
In this section we document the criteria that guided the anglicism selection, writing process and punctuation transformations behind the examples in BLAS.

\subsection{Anglicism selection criteria}
\label{appendix:selection}
The first step when writing the examples for BLAS was to select the anglicisms that would serve to evaluate a given linguistic attribute. 
Anglicisms were selected by hand from the anglicism dictionary \textit{Nuevo diccionario de anglicismos} by \citet{gonzalez2009nuevo} and from Observatorio Lázaro, an online resource that automatically compiles anglicisms used in the Spanish media \citep{observatorio}.

The following criteria were taken into account for the selection:

\paragraph{Frequency criteria}
With BLAS we wanted to evaluate how well models perform when they face anglicisms that are not seen in training. 
In other words, we wanted to ensure that most of the borrowings in BLAS would be unseen. 
However BLAS comes with no associated training data and, being a test-only dataset, we cannot anticipate or assume the type of data that future models evaluated on BLAS will be trained on. 
This means that we cannot guarantee that the anglicisms in BLAS will be unseen for the models.

To mitigate this limitation, we made a conscious effort to include borrowings in BLAS that are more niche and less widespread. 
Consequently, during the selection process we prioritized less obvious anglicisms over anglicisms that we would expect to find in a monolingual dictionary of Spanish, or that are so frequent that they could be expected to appear in a generalistic training set (while still keeping a balance in terms of domains represented). 

We also consciously avoided extremely rare borrowings and nonce borrowings; to do so, we made sure that the selected borrowings appeared in the Spanish general media and newspapers. 
For borrowings retrieved from Observatorio Lázaro, this means that we looked for borrowings that had been documented more than 10 times and less than 50 in Spanish newspapers.

\paragraph{Sentence position criteria}
Because we wanted to assess performance on different sentence positions, we prioritized whenever possible to select borrowings that could work equally well mid sentence and in sentence initial position.

\paragraph{Topic and shape variety} The selection was also curated to ensure that a variety of topics and shapes were covered, so that not all anglicisms belonged to the same domain or had similar character combinations.

\paragraph{Unassimilation criteria}
Our aim with BLAS is that it can serve as a stable benchmark that future models can be evaluated on.
This wish clashes against a linguistic fact of life: the fact that a borrowing that is frequent enough may very easily end up being assimilated into the recipient language and thus not perceived as a borrowing anymore.
This poses the risk of including borrowings in BLAS that will eventually become assimilated and will not serve our purpose anymore. 
This is particularly true for graphotactically compliant anglicisms (that is, anglicisms whose spelling does not violate the spelling rules of Spanish), as the fact of being compliant may cause that speakers rapidly forget that the word was borrowed, thus facilitating its assimilation process (while non-compliance preserves the speakers' perception of word being foreign).

To account for this, for compliant borrowings we chose words where, even if their spelling did not violate the graphotactics of Spanish, there was no correspondence between the spelling of the word and its pronunciation according to the Spanish pronunciation expectations.
In other words, a word like \textit{barman} is a graphotactically compliant borrowing whose spelling matches the expectations of how the word is pronounced in Spanish  (\textipa{/"baR.man/}). 
This makes the borrowing \textit{barman} a bad choice for BLAS, because due to its shape and pronunciation it makes it a good candidate for rapid assimilation. 
On the other hand, borrowings like \textit{paper} (\textipa{/"peI.peR/}) or \textit{eyeliner} (\textipa{/aI."laI.neR/}), while still being fully compliant in their spelling, are currently pronounced in a way that violates the expectations of Spanish, which ensures that they are perceived as foreign by native speakers and that are far from becoming assimilated in the near future.
Consequently, for graphotactically compliant borrowings we chose borrowings where there was no correspondence between its spelling and pronunciation.

\subsubsection{Compliant and not-compliant spans}
\label{benchmark_A}
Compliant anglicisms are anglicisms whose spelling complies with the Spanish rules of spelling. 
On the other hand, non-compliant anglicisms are anglicisms whose spelling violates the rules of spelling. 
Non-compliant borrowings are easier to spot, because their spelling violates the expectations that Spanish speakers have about how a word should be written.
As a result, previous work had already pointed out that models tended to privilege non-compliant borrowings over compliant ones \citep{mellado2024characterizing}.
To assess differences in performance between compliant and non-compliant borrowings, we selected the following anglicisms: 
\begin{itemize}
\item[-] 20 graphotactically compliant anglicisms (10 single word anglicisms, 10 multiword anglicisms): for example, \textit{pipeline}, \textit{medal race}.
\item[-] 20 graphotactically non-compliant anglicisms (10 single word anglicisms, 10 multiword anglicisms): for example, \textit{cliffhanger}, \textit{fact checker}.
\item[-] 10 graphotactically mixed-compliant anglicisms, where at least one token is compliant and one token is non-compliant (all multiword): for example, \textit{tie break}.
\end{itemize}


\subsubsection{Ambiguous spans}
\label{benchmark_B}
Ambiguous anglicisms are anglicisms that also exist as fully Spanish words (such as \textit{pie}, that is a native word when it means ``foot'' but a borrowing pronounced as \textipa{/paI/} when it refers to a dish). 
Ambiguous spans can easily confuse models, as they can be seen during training as native words or found as native words in Spanish lexicons. 
Our aim was to evaluate specifically how well models perform on borrowings that are prone to label mismatch. 

As a result, we collected a set of anglicisms that are ambiguous with Spanish words (either single word items or multiword items), following the same procedure we described in section \ref{benchmark_A}. 
The following borrowings were considered: 

\begin{itemize}
\item[-] 3 single-word ambiguous borrowings: for example, \textit{pie}.
\item[-] 10 multiword ambiguous borrowings: for example, \textit{global director}.
\item[-] 10 multiword mixed-ambiguous borrowings (where at least one of the words is ambiguous and one of the tokens is not): for example, \textit{escape room}.
\end{itemize}

We identified very few candidates that were single-word ambiguous (only 3), but we still considered them to be interesting enough to include them in BLAS, as these examples can help us assess models' performance on an extremely difficult case.

For pure ambiguous borrowings (that is, not mixed ambiguous), we lifted the restriction that borrowings had to work in sentence initial position: given the limited number of ambiguous borrowings and in order to avoid populating our benchmark with unnatural examples (not all borrowings work equally well on sentence initial position and how natural a borrowing in sentence initial position sounds is given by the meaning of the borrowing), we focused exclusively on mid sentence position examples, where all borrowings work equally well.

Consequently, for pure ambiguous borrowings we only wrote one example per selected borrowing, in which the borrowing appeared mid-sentence.
We still wrote two examples (one sentence initial, one mid-sentence) for mixed-ambiguous anglicisms, as mixed-ambiguous are not rare and we could easily find good candidates for mixed-ambiguous that could work on sentence initial position.

\subsubsection{Adjacent spans}
\label{benchmark_C}
Finally, we wanted our benchmark to evaluate performance over sentences with adjacent borrowings, ie. two independent borrowings that are collocated next to one another in the sentence (as in \textit{look total black}), as this type of sentences are prone to segmentation errors.
We selected the following 20 pairs of borrowings that worked well collocated: 
\begin{itemize}
    \item[-] 10 pairs of collocated borrowings, where both borrowings were single-word borrowings; for example, \textit{marketing online} in \textit{Una campaña de marketing online}\footnote{``An online marketing campaign.''}.
    \item[-] 10 pairs of collocated borrowings, where at least one of the borrowings in the pair was multiword; for example, \textit{look total black} in \textit{La actriz lució un look total black}\footnote{``The actress wore a total black look.''}.
\end{itemize}

Given that not all borrowings will work equally well collocated with other borrowings and in order to ensure natural sounding sentences, the selection of borrowings in this case was more lenient: more frequent borrowings were allowed in the selection and only mid-sentence position sentences were included, as what we were interested in assessing with these examples was not the retrieval capability of the models but their ability to correctly segment borrowings that may have been previously seen.

Similarly to ambiguous anglicisms, for adjacent anglicisms we only wrote one example per selected pair of borrowings, in which the borrowings appeared mid-sentence.

\begin{figure*}[t]
\includegraphics[width=\textwidth]{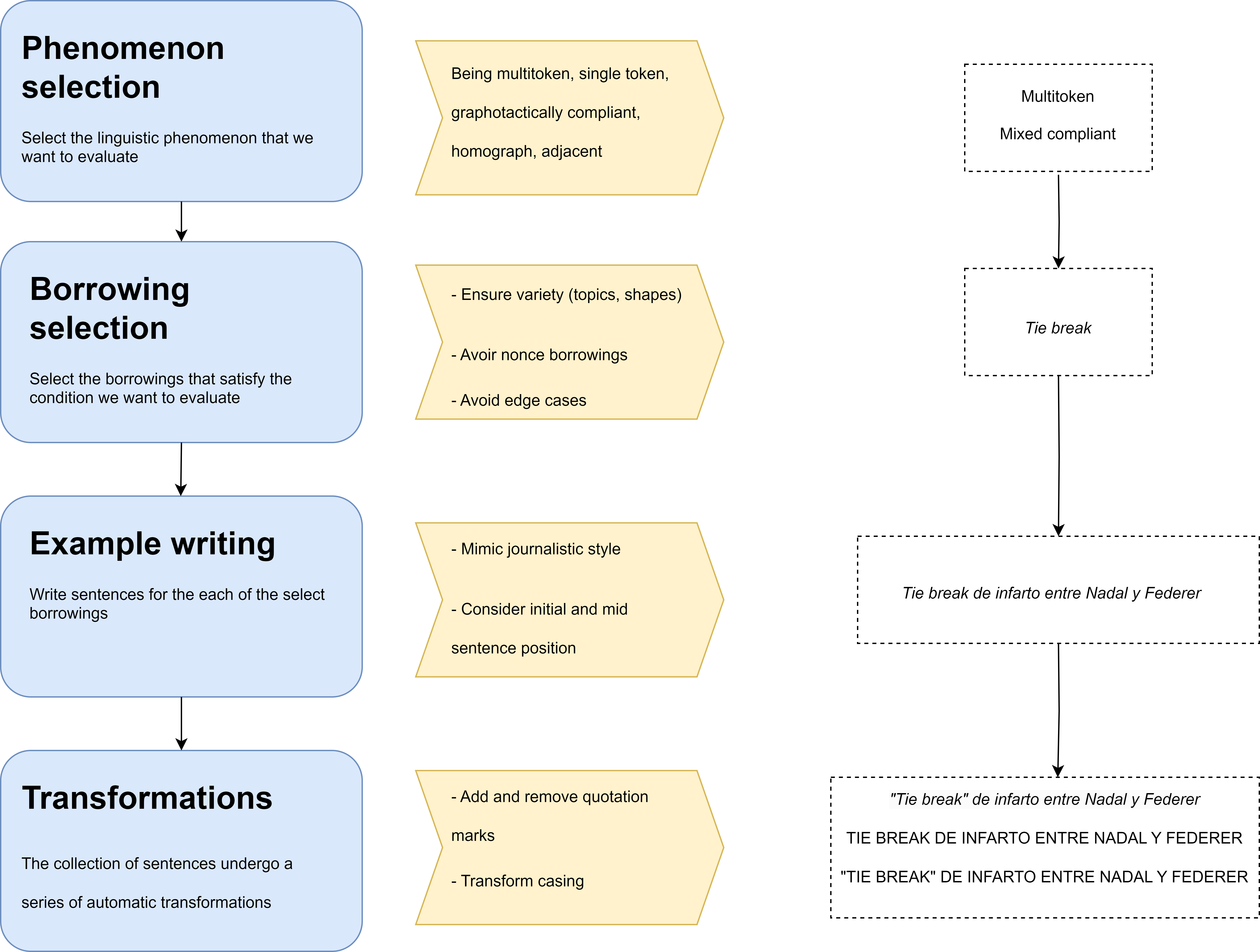}
\caption{Creation process of the examples in BLAS.}
\label{fig:blas}
\end{figure*}

\subsection{Writing process}
\label{appendix:writing}
BLAS was populated with sentences that contained the anglicisms selected in Appendix \ref{appendix:selection}.

Sentences in BLAS were created ad hoc by hand by a linguist inspired on real examples from journalistic sources.
Sentences in BLAS are characterized by a given configuration of linguistic attributes: a group of sentences contains multiword anglicisms that are non-compliant and that appear mid-sentence, another group of sentences contain the same multiword anglicisms but in sentence initial position, etc.

All examples in BLAS are original and were written from scratch, although some of them were inspired by real world examples and all of them try to mimic the journalistic style of the Spanish newspapers and media.
The linguist that authored the examples was a native speaker of Spanish with experience in anglicism annotation and journalistic writing.
Figure \ref{fig:blas} reflects the writing process behind the examples in BLAS. Table \ref{tab:sentence_num} reflects number of sentences per linguistic attribute.

\subsection{Punctuation transformations}
\label{appendix:transformations}
The process described in Appendix \ref{appendix:writing} yielded 153 sentences. These sentences underwent a series of automatic punctuation transformations.
We considered two types of punctuation transformations: casing and quotation marks.

For casing, the following transformations were performed:\\
    \noindent -- Standard casing (no transformations, text as is).\\
    -- all words are lowercase.\\
    -- ALL WORDS ARE UPPERCASE.\\
    -- All Words Are Capitalized.\\
    -- Only BORROWINGS are in uppercase.\\
    -- Only Borrowings are capitalized.\\
For quotation marks, the following perturbations were performed:

    \noindent -- All ``anglicisms'' appear between quotation marks.\\
    -- No anglicism is written with quotation marks.

This amounts to a total of twelve different configurations (six casing transformations with quotation marks, six without them). 
Each sentence in the original collection of 153 sentences underwent these twelve transformations.
Table \ref{tab:transformations} shows an example of all twelve casing and punctuation transformations for a single sentence.

\section{Prompting methodology with Llama3}
\label{appendix:llama3}
Our prompting methodology was inspired by the work of \citet{ashok_promptner_2023}, which offer prompted-based heuristics to apply to generative LLMs for the task of NER.
Their experiments show that in addition to the few-shot examples, providing the model with a modular definition of the relevant entity types in the prompt produces better results than pure few-shot prompting.


To that end, we prompted 8B-Llama3 \citep{grattafiori2024llama3herdmodels,llama3modelcard} with a definition of what an anglicism is and a description of the task the model was expected to complete. 
The definition included a list of four examples of decontextualized anglicisms (\textit{online}, \textit{machine learning}, \textit{podcast}, \textit{blockchain}) and a description of some of the characteristics and contexts an anglicism may appear in (single-word, multiword, adjacent, quoted or unquoted) illustrated by a minimal collection of the following three examples:
\begin{itemize}
    \item \textit{El evento se retransmitirá por \underline{streaming}}.\footnote{``The event will be  broadcast live via streaming''.}
    \item \textit{La marca ha sacado a la venta un nuevo \underline{concept car}}. \footnote{``The company has released a new concept car''.}
    \item \textit{Compré mi \underline{smartphone} \underline{online}}.\footnote{``I bought my smartphone online''.}
\end{itemize}

The definition explicitly mentioned that proper names such as entity names should not be considered anglicisms.

Figure \ref{fig:prompt} displays the prompt we used, which draws from the style of prompting for NER proposed by \citeauthor{ashok_promptner_2023}.

We chose Llama3 for our experiments because at the time of conducting our experiments Llama3 was among the best performing models while being open source, which ensured that our conclusions could be tested, replicated and easily transferred to other languages. 
The model we used was the small Llama3 (8B parameter), as it is less resource intensive than larger versions.
We queried it through \texttt{ollama} library\footnote{\url{https://ollama.com/}}, with default context window size of 2048 and temperature set to 0.0.

\section{Prediction experiments}
\label{appendix:spans_COALAS}

Tables \ref{tab:spans_coalas}, \ref{tab:spans_coalas_dev}, \ref{tab:spans_calcs_dev}, \ref{tab:spans_calcs_testA} and \ref{tab:spans_calcs_testB} show number of spans per type on each of the five external datasets: COALAS test set and development set \citep{alvarez-mellado-lignos-2022-detecting} and CALCS development and test sets A and B \citep{alvarez-mellado-2020-annotated}.

Table \ref{tab:correl} displays the predicted recall scores and ranking, true recall scores and ranking and true F1 scores and ranking for all six models over the five external datasets, along with Pearson and Spearman correlation coefficients.

Table \ref{tab:avg_correl} displays average and median correlation coefficients between the predicted results and true results, along with correlation coefficients between BLAS results and true results.

\begin{table*}[]
\tiny
\centering
\resizebox{\textwidth}{!}{%
\begin{tabular}{@{}lllllr@{}}
\toprule
\textbf{Casing}     & \textbf{Quotation marks} & \textbf{Type}   & \textbf{Length} & \textbf{Position} & \textbf{Number} \\ \midrule
Standard            & w/o quot.                & non compliant   & single          & mid               & 688             \\
Standard            & w/o quot.                & compliant       & single          & mid               & 141             \\
Standard            & w/o quot.                & non compliant   & multi           & mid               & 123             \\
Standard            & w/o quot.                & mixed compliant & multi           & mid               & 107             \\
Standard            & with quot.               & non compliant   & single          & mid               & 42              \\
Standard            & with quot.               & non compliant   & multi           & mid               & 20              \\
Standard            & with quot.               & mixed compliant & multi           & mid               & 20              \\
Standard            & w/o quot.                & adjacent        & single          & mid               & 14              \\
Standard            & w/o quot.                & mixed ambiguous & multi           & mid               & 13              \\
Standard            & w/o quot.                & non compliant   & single          & ini               & 11              \\
Standard            & w/o quot.                & compliant       & multi           & mid               & 9               \\
Standard            & with quot.               & compliant       & multi           & mid               & 9               \\
Spans are titlecase & w/o quot.                & non compliant   & single          & mid               & 8               \\
Standard            & with quot.               & compliant       & single          & mid               & 7               \\
Standard            & w/o quot.                & non compliant   & multi           & ini               & 4               \\
Text is lowercase   & w/o quot.                & non compliant   & multi           & mid               & 3               \\
Standard            & w/o quot.                & ambiguous       & multi           & mid               & 3               \\
Standard            & w/o quot.                & mixed compliant & multi           & ini               & 3               \\
Spans are titlecase & w/o quot.                & compliant       & single          & mid               & 2               \\
Spans are titlecase & w/o quot.                & mixed ambiguous & multi           & mid               & 2               \\
Standard            & w/o quot.                & ambiguous       & single          & mid               & 2               \\
Standard            & w/o quot.                & adjacent        & multi           & mid               & 2               \\
Spans are uppercase & w/o quot.                & non compliant   & single          & ini               & 1               \\
Spans are titlecase & w/o quot.                & non compliant   & multi           & mid               & 1               \\
Spans are titlecase & with quot.               & non compliant   & multi           & mid               & 1               \\
Standard            & w/o quot.                & compliant       & single          & ini               & 1               \\
Standard            & w/o quot.                & mixed ambiguous & multi           & ini               & 1               \\
Spans are titlecase & w/o quot.                & mixed compliant & multi           & mid               & 1               \\ \bottomrule
\end{tabular}%
}
\caption{Number of spans per type in COALAS test set.}
\label{tab:spans_coalas}
\end{table*}

\begin{table*}[]
\tiny
\centering
\resizebox{\textwidth}{!}{%
\begin{tabular}{@{}lllllr@{}}
\toprule
\textbf{Casing}     & \textbf{Quotation marks} & \textbf{Type}   & \textbf{Length} & \textbf{Position} & \textbf{Number} \\ \midrule
Standard casing  &  with quot.   & non compliant   & single & mid & 90                   \\
Standard casing  & w/o quot. & non compliant   & single & mid & 79                   \\
Standard casing  & with quot.   & non compliant   & multi  & mid & 30                   \\
Standard casing  & with quot.   & mixed compliant & multi  & mid & 22                   \\
Standard casing  & with quot.   & compliant       & single & mid & 19                   \\
Standard casing  & w/o quot. & compliant       & single & mid & 17                   \\
Standard casing  & with quot.   & compliant       & multi  & mid & 9                    \\
Standard casing  & w/o quot. & non compliant   & multi  & mid & 8                    \\
Standard casing  & w/o quot. & mixed compliant & multi  & mid & 6                    \\
Spans are titlecase & with quot.   & mixed compliant & multi  & mid & 5                    \\
Spans are titlecase & with quot.   & non compliant   & single & mid & 4                    \\
Spans are titlecase & w/o quot. & mixed compliant & multi  & mid & 3                    \\
Standard casing  & with quot.   & non compliant   & single & ini & 2                    \\
Spans are titlecase & w/o quot. & non compliant   & multi  & mid & 2                    \\
Standard casing  & w/o quot. & compliant       & multi  & mid & 2                    \\
Standard casing  & w/o quot. & non compliant   & single & ini & 1                    \\
Spans are titlecase & with quot.   & non compliant   & multi  & mid & 1                    \\
Standard casing  & w/o quot. & non compliant   & multi  & ini & 1                    \\
Standard casing  & with quot.   & compliant       & single & ini & 1                    \\
Standard casing  & w/o quot. & compliant       & single & ini & 1                    \\
Standard casing  & with quot.   & ambiguous       & multi  & mid & 1                    \\
Standard casing  & with quot.   & ambiguous       & single & mid & 1                    \\
Spans are titlecase & with quot.   & mixed ambiguous & multi  & mid & 1                                \\ \bottomrule
\end{tabular}%
}
\caption{Number of spans per type in COALAS development set.}
\label{tab:spans_coalas_dev}
\end{table*}

\begin{table*}[]
\tiny
\centering
\resizebox{\textwidth}{!}{%
\begin{tabular}{@{}lllllr@{}}
\toprule
\textbf{Casing}     & \textbf{Quotation marks} & \textbf{Type}   & \textbf{Length} & \textbf{Position} & \textbf{Number} \\ \midrule
Standard casing  & w/o quot. & non compliant   & single & mid & 67 \\
Standard casing  & with quot.   & non compliant   & single & mid & 55 \\
Standard casing  & w/o quot. & compliant       & single & mid & 18 \\
Standard casing  & with quot.   & compliant       & single & mid & 15 \\
Standard casing  & w/o quot. & non compliant   & single & ini & 10 \\
Standard casing  & w/o quot. & non compliant   & multi  & mid & 9  \\
Standard casing  & with quot.   & non compliant   & multi  & mid & 8  \\
Standard casing  & with quot.   & mixed compliant & multi  & mid & 7  \\
Standard casing  & w/o quot. & ambiguous       & multi  & mid & 5  \\
Spans are titlecase & w/o quot. & non compliant   & single & mid & 3  \\
Standard casing  & with quot.   & compliant       & multi  & mid & 3  \\
Spans are titlecase & w/o quot. & mixed compliant & multi  & mid & 3  \\
Standard casing  & w/o quot. & mixed compliant & multi  & mid & 3  \\
Standard casing  & w/o quot. & mixed ambiguous & multi  & mid & 3  \\
Standard casing  & with quot.   & mixed ambiguous & multi  & mid & 2  \\
Standard casing  & with quot.   & non compliant   & single & ini & 1  \\
Text is lowercase  & w/o quot. & non compliant   & single & mid & 1  \\
Spans are titlecase & w/o quot. & compliant       & single & mid & 1  \\
Standard casing  & with quot.   & compliant       & single & ini & 1  \\
Text is titlecase  & w/o quot. & compliant       & single & ini & 1  \\
Standard casing  & with quot.   & ambiguous       & multi  & mid & 1  \\
Standard casing  & w/o quot. & mixed compliant & multi  & ini & 1  \\
Standard casing  & with quot.   & mixed ambiguous & multi  & ini & 1                               \\ \bottomrule
\end{tabular}%
}
\caption{Number of spans per type in CALCS development set.}
\label{tab:spans_calcs_dev}
\end{table*}

\begin{table*}[]
\tiny
\centering
\resizebox{\textwidth}{!}{%
\begin{tabular}{@{}lllllr@{}}
\toprule
\textbf{Casing}     & \textbf{Quotation marks} & \textbf{Type}   & \textbf{Length} & \textbf{Position} & \textbf{Number} \\ \midrule
Standard casing  & w/o quot. & non compliant   & single & mid & 77 \\
Standard casing  & with quot.   & non compliant   & single & mid & 47 \\
Standard casing  & w/o quot. & compliant       & single & mid & 19 \\
Standard casing  & with quot.   & compliant       & single & mid & 14 \\
Standard casing  & w/o quot. & non compliant   & multi  & mid & 7  \\
Standard casing  & w/o quot. & mixed compliant & multi  & mid & 6  \\
Standard casing  & with quot.   & mixed compliant & multi  & mid & 6  \\
Standard casing  & with quot.   & non compliant   & multi  & mid & 5  \\
Standard casing  & w/o quot. & non compliant   & single & ini & 4  \\
Standard casing  & with quot.   & compliant       & multi  & mid & 4  \\
Standard casing  & with quot.   & mixed compliant & multi  & ini & 2  \\
Spans are titlecase & with quot.   & mixed compliant & multi  & mid & 2  \\
Spans are titlecase & w/o quot. & non compliant   & single & mid & 2  \\
Standard casing  & w/o quot. & mixed ambiguous & multi  & mid & 2  \\
Standard casing  & w/o quot. & mixed compliant & multi  & ini & 1  \\
Spans are titlecase & w/o quot. & mixed compliant & multi  & mid & 1  \\
Spans are titlecase & with quot.   & non compliant   & single & mid & 1  \\
Text is lowercase  & with quot.   & non compliant   & single & mid & 1  \\
Standard casing  & with quot.   & non compliant   & single & ini & 1  \\
Standard casing  & w/o quot. & non compliant   & multi  & ini & 1  \\
Standard casing  & w/o quot. & compliant       & multi  & mid & 1  \\
Standard casing  & w/o quot. & compliant       & multi  & ini & 1  \\
Text is lowercase & w/o quot. & compliant       & single & mid & 1  \\
Standard casing  & with quot.   & compliant       & single & ini & 1  \\
Standard casing  & w/o quot. & compliant       & single & ini & 1  \\
Spans are titlecase & w/o quot. & mixed ambiguous & multi  & mid & 1  \\
Standard casing  & with quot.   & mixed ambiguous & multi  & mid & 1  \\
Spans are titlecase & with quot.   & mixed ambiguous & multi  & mid & 1  \\
Standard casing  & w/o quot. & ambiguous       & multi  & mid & 1                               \\ \bottomrule
\end{tabular}%
}
\caption{Number of spans per type in CALCS test set A.}
\label{tab:spans_calcs_testA}
\end{table*}

\begin{table*}[]
\tiny
\centering
\resizebox{\textwidth}{!}{%
\begin{tabular}{@{}lllllr@{}}
\toprule
\textbf{Casing}     & \textbf{Quotation marks} & \textbf{Type}   & \textbf{Length} & \textbf{Position} & \textbf{Number} \\ \midrule
Standard casing  & w/o quot. & non compliant   & single & mid & 43                   \\
Standard casing  & w/o quot. & compliant       & single & mid & 25                   \\
Standard casing  & with quot.   & non compliant   & single & mid & 20                   \\
Standard casing  & with quot.   & compliant       & single & mid & 12                   \\
Standard casing  & with quot.   & mixed compliant & multi  & mid & 4                    \\
Standard casing  & w/o quot. & non compliant   & multi  & mid & 3                    \\
Standard casing  & with quot.   & non compliant   & multi  & mid & 3                    \\
Standard casing  & w/o quot. & non compliant   & single & ini & 2                    \\
Standard casing  & with quot.   & non compliant   & single & ini & 2                    \\
Standard casing  & w/o quot. & mixed compliant & multi  & mid & 2                    \\
Standard casing  & w/o quot. & ambiguous       & multi  & mid & 2                    \\
Standard casing  & w/o quot. & compliant       & single & ini & 1                    \\
Standard casing  & with quot.   & compliant       & multi  & mid & 1                    \\
Spans are titlecase & with quot.   & compliant       & multi  & mid & 1                    \\
Standard casing  & w/o quot. & compliant       & multi  & mid & 1                    \\
Spans are titlecase & with quot.   & non compliant   & single & mid & 1                    \\
titlecased text  & with quot.   & non compliant   & single & ini & 1                    \\
Spans are titlecase & w/o quot. & mixed compliant & multi  & mid & 1                    \\
Standard casing  & with quot.   & mixed ambiguous & multi  & mid & 1                    \\                        \bottomrule
\end{tabular}%
}
\caption{Number of spans per type in CALCS test set B.}
\label{tab:spans_calcs_testB}
\end{table*}

\begin{figure*}[h] 
\color{blue}
\texttt{Un anglicismo es un préstamo crudo incorporado del inglés que se usa dentro de una frase en castellano sin adaptación, palabras como `online', `machine learning', `podcast' o `blockchain'. Un anglicismo puede estar formado por una única palabra (como `streaming' en `El evento se retransmitirá por streaming'), por varias palabras (como en `La marca ha sacado a la venta un nuevo concept car') o ser dos anglicismos independientes que aparecen uno al lado del otro (como `smartphone' y `online' en `Compré mi smartphone online'), y pueden aparecer entrecomillados o no. Los nombres propios (como los nombres de personas, lugares o títulos de obras de ficción) no son anglicismos."} 
\linebreak
\linebreak
\color{red}
\texttt{Voy a darte una frase en castellano, tu misión es identificar si la frase contiene algún anglicismo. La frase puede contener un anglicismo, varios o no contener ninguno.  Responde solo con el segmento que cumpla la condición de ser anglicismo (si es que lo hay) y sin añadir ninguna consideración más. Si la frase no contiene ningún anglicismo, responde 'None'. Si la frase contiene más de 1 anglicismo, devuelve todos los anglicismos, con un anglicismo por línea (es decir, añade un salto de línea entre anglicismos). No alucines ni añadas palabras que no estén en la frase original. Esta es la frase: \\\\El problema de la  "fast fashion" es triple : consume recursos desmesuradamente , produce toneladas de basura de difícil reciclaje y fomenta la explotación laboral de la industria textil .}
\linebreak
\linebreak
\color{orange}
\texttt{"fast fashion"}
\color{black}
\caption{Prompt provided to Llama3 inspired by \citeauthor{ashok_promptner_2023}'s template: \color{blue}{Definition of the task with examples}, \color{red}{question and task}, \color{orange}{response by Llama3}\color{black}{.\\
Translation:} \color{blue}{An anglicism is a lexical borrowing that is incorporated from English into Spanish without adaptation, such as `online', `machine learning', `podcast' or `blockchain'. An anglicism can be composed of one single word (like `streaming' in `El evento se retransmitirá por streaming'), several words (like `concept car' in `La marca
ha sacado a la venta un nuevo concept car') or two adjacent anglicisms (like  `smartphone' y `online' en  `Compré mi smartphone online') and can be written between quotation marks or not. Proper foreign names (like person names, location names or titles) are not anglicisms.}\\
\color{red}{You will be given a sentence in Spanish, your mission is identifying if the sentence contains anglicisms. The sentence may contain one anglicism, several anglicisms or none. Reply only with the span that satisfies the requirement of being an anglicism (if there is any), without any other information. If the sentence contains no anglicism simply reply `None'. If the sentence contains more than one anglicism, reply with one anglicism per line (leave a linebreak between anglicisms). This is the sentence:\\
El problema de la "fast fashion" es triple : consume recursos
desmesuradamente , produce toneladas de basura de difícil reciclaje
y fomenta la explotación laboral de la industria textil .}\\
\color{orange}{"fast fashion"}}
\label{fig:prompt}
\end{figure*}

\begin{landscape}
\begin{table}[]
\resizebox{\columnwidth}{!}{%
\begin{tabular}{@{}l|r|rrrrrr|rrrrrr|rrrrrr|rrrrrr|rrrrrr|@{}}
\toprule
\multicolumn{1}{l|}{\multirow{2}{*}{\textbf{Models}}} & \multicolumn{1}{l|}{\textbf{BLAS}} & \multicolumn{6}{c|}{\textbf{COALAS (test set)}}                                                                                                                                           & \multicolumn{6}{c|}{\textbf{COALAS (dev set)}}                                                                                                                                            & \multicolumn{6}{c|}{\textbf{CALCS (test set A)}}                                                                                                                                          & \multicolumn{6}{c|}{\textbf{CALCS (test set B)}}                                                                                                                                          & \multicolumn{6}{c|}{\textbf{CALCS (dev set)}}                                                                                                                                             \\ \cmidrule(l){2-32} 
\multicolumn{1}{c|}{}                                 & \multicolumn{1}{c|}{\textbf{R}}    & \multicolumn{1}{l}{\textbf{Pred. R}} & \multicolumn{1}{l}{} & \multicolumn{1}{l}{\textbf{True R}} & \multicolumn{1}{l}{} & \multicolumn{1}{l}{\textbf{True F1}} & \multicolumn{1}{l|}{} & \multicolumn{1}{l}{\textbf{Pred. R}} & \multicolumn{1}{l}{} & \multicolumn{1}{l}{\textbf{True R}} & \multicolumn{1}{l}{} & \multicolumn{1}{l}{\textbf{True F1}} & \multicolumn{1}{l|}{} & \multicolumn{1}{l}{\textbf{Pred. R}} & \multicolumn{1}{l}{} & \multicolumn{1}{l}{\textbf{True R}} & \multicolumn{1}{l}{} & \multicolumn{1}{l}{\textbf{True F1}} & \multicolumn{1}{l|}{} & \multicolumn{1}{l}{\textbf{Pred. R}} & \multicolumn{1}{l}{} & \multicolumn{1}{l}{\textbf{True R}} & \multicolumn{1}{l}{} & \multicolumn{1}{l}{\textbf{True F1}} & \multicolumn{1}{l|}{} & \multicolumn{1}{l}{\textbf{Pred. R}} & \multicolumn{1}{l}{} & \multicolumn{1}{l}{\textbf{True R}} & \multicolumn{1}{l}{} & \multicolumn{1}{l}{\textbf{True F1}} & \multicolumn{1}{l|}{} \\ \midrule
\textbf{CRF}                                          & 6.50                               & 26.58                                  & \#6                   & 44.31                               & \#5                   & 56.54                                & \#5                    & 33.13                                  & \#6                   & 68.63                               & \#5                   & 71.31                                & \#5                    & 29.00                                  & \#6                   & 65.57                               & \#5                   & 70.20                                & \#5                    & 24.76                                  & \#6                   & 53.17                               & \#5                   & 61.19                                & \#5                    & 28.99                                  & \#6                   & 64.84                               & \#5                   & 71.90                                & \#5                    \\
\textbf{BETO}                                         & 23.55                              & 86.87                                  & \#3                   & 77.99                               & \#4                   & 82.36                                & \#4                    & 87.62                                  & \#3                   & 84.05                               & \#4                   & 78.81                                & \#4                    & 83.58                                  & \#4                   & 83.49                               & \#3                   & 82.71                                & \#4                    & 79.13                                  & \#4                   & 81.75                               & \#4                   & 76.87                                & \#3                    & 81.60                                  & \#4                   & 86.30                               & \#1                   & 87.30                                & \#2                    \\
\textbf{mBERT}                                        & 23.80                              & 86.86                                  & \#4                   & 78.85                               & \#3                   & 83.64                                & \#3                    & 87.10                                  & \#4                   & 84.31                               & \#3                   & 82.21                                & \#3                    & 83.86                                  & \#3                   & 85.85                               & \#1                   & 84.26                                & \#2                    & 83.17                                  & \#3                   & 84.13                               & \#3                   & 76.26                                & \#4                    & 82.73                                  & \#3                   & 86.30                               & \#1                   & 86.30                                & \#3                    \\
\textbf{BiLSTM (unad)}                                & 23.55                              & 87.86                                  & \#2                   & 80.88                               & \#2                   & 85.14                                & \#1                    & 88.73                                  & \#2                   & 89.05                               & \#1                   & 86.73                                & \#1                    & 84.85                                  & \#2                   & 82.08                               & \#4                   & 83.86                                & \#3                    & 84.04                                  & \#2                   & 84.92                               & \#2                   & 84.25                                & \#1                    & 83.06                                  & \#2                   & 83.11                               & \#4                   & 85.25                                & \#4                    \\
\textbf{BiLSTM (cs)}                                  & 23.75                              & 92.96                                  & \#1                   & 85.76                               & \#1                   & 84.22                                & \#2                    & 91.08                                  & \#1                   & 86.57                               & \#2                   & 84.66                                & \#2                    & 89.43                                  & \#1                   & 83.96                               & \#2                   & 84.96                                & \#1                    & 89.52                                  & \#1                   & 85.71                               & \#1                   & 83.08                                & \#2                    & 87.53                                  & \#1                   & 85.84                               & \#3                   & 87.44                                & \#1                    \\
\textbf{Llama3}                                       & 36.37                              & 42.93                                  & \#5                   & 33.33                               & \#6                   & 39.73                                & \#6                    & 55.46                                  & \#5                   & 53.27                               & \#6                   & 40.30                                & \#6                    & 48.53                                  & \#5                   & 53.30                               & \#6                   & 29.62                                & \#6                    & 43.17                                  & \#5                   & 40.48                               & \#6                   & 17.17                                & \#6                    & 48.49                                  & \#5                   & 50.68                               & \#6                   & 30.41                                & \#6                    \\ \midrule
\textbf{Correlation}                                  & \multicolumn{1}{l|}{}              & \multicolumn{1}{l}{}                   & \multicolumn{1}{l}{} & 0.94                                & 0.89                 & 0.89                                 & 0.83                  & \multicolumn{1}{l}{}                   & \multicolumn{1}{l}{} & 0.79                                & 0.83                 & 0.61                                 & 0.83                  & \multicolumn{1}{l}{}                   & \multicolumn{1}{l}{} & 0.85                                & 0.66                 & 0.64                                 & 0.89                  & \multicolumn{1}{l}{}                   & \multicolumn{1}{l}{} & 0.91                                & 0.94                 & 0.69                                 & 0.83                  & \multicolumn{1}{l}{}                   & \multicolumn{1}{l}{} & 0.84                                & 0.41                 & 0.63                                 & 0.71                  \\ \bottomrule
\end{tabular}%
}
\caption{Pearson correlation coefficient over true recall scores, F1 scores and ranking obtained by six models over different datasets and the predicted recall scores and rankings obtained by extrapolating BLAS scores.}
\label{tab:correl}
\end{table}

\begin{table}[]
\resizebox{\textwidth}{!}{%
\begin{tabular}{@{}lrrrr@{}}
\toprule
\textbf{Dataset} & \multicolumn{1}{l}{\textbf{\begin{tabular}[c]{@{}l@{}}Recall correl. with\\ BLAS prediction\end{tabular}}} & \multicolumn{1}{l}{\textbf{\begin{tabular}[c]{@{}l@{}}Ranking correl. with\\ BLAS prediction\end{tabular}}} & \multicolumn{1}{l}{\textbf{\begin{tabular}[c]{@{}l@{}}Recall correl. \\ with BLAS\end{tabular}}} & \multicolumn{1}{l}{\textbf{\begin{tabular}[c]{@{}l@{}}Ranking correl. \\ with BLAS\end{tabular}}} \\ \midrule
COALAS test set  & 0.94                                                                                                  & 0.89                                                                                                   & -0.04                                                                                            & -0.08                                                                                             \\
COALAS dev set   & 0.79                                                                                                  & 0.83                                                                                                   & -0.23                                                                                            & -0.17                                                                                             \\
CALCS test set A & 0.85                                                                                                  & 0.66                                                                                                   & -0.19                                                                                            & -0.03                                                                                             \\
CALCS test set B & 0.91                                                                                                   & 0.94                                                                                                   & -0.10                                                                                             & -0.37                                                                                             \\
CALCS dev set    & 0.83                                                                                                  & 0.43                                                                                                   & -0.20                                                                                             & 0.06                                                                                              \\
Average correl.  & \textbf{0.86}                                                                                        & \textbf{0.75}                                                                                         & \textbf{-0.15}                                                                                  & \textbf{-0.12}                                                                                   \\
Median correl.   & \textbf{0.85}                                                                                         & \textbf{0.83}                                                                                          & \textbf{-0.19}                                                                                   & \textbf{-0.08}                                                                                    \\ \bottomrule
\end{tabular}%
}
\caption{Pearson and Spearman correlation coefficients obtained from comparing recall score distributions and ranking between the true results obtained on a dataset, the predicted results obtained by extrapolating BLAS scores and BLAS results.}
\label{tab:avg_correl}
\end{table}

\end{landscape}

\end{document}